\pdfoutput=1

\documentclass[11pt]{article}

\usepackage[final]{acl}

\usepackage{times}
\usepackage{latexsym}
\usepackage{makecell}
\usepackage{booktabs}
\usepackage{breqn}

\usepackage{geometry}
\usepackage{graphicx} 
\usepackage{amsmath}  
\usepackage{xcolor}
\usepackage{float}
\usepackage[T1]{fontenc}

\usepackage[utf8]{inputenc}
\usepackage{multirow}

\usepackage{microtype}

\usepackage{inconsolata}

\usepackage{graphicx}

\usepackage{amsmath}
\usepackage{amsfonts} 

\usepackage{autobreak}

\usepackage{caption}
\usepackage[normalem]{ulem}
\usepackage{subcaption}

\usepackage{booktabs}
\usepackage{multirow}
\usepackage{xcolor, colortbl} 
\usepackage{array}            
\usepackage{caption}          
\usepackage{soul}
\usepackage{hyperref}
\definecolor{lightgray}{gray}{0.95}
\definecolor{lightblue}{RGB}{175, 210, 230}
\definecolor{lightorange}{RGB}{255, 210, 170}
\definecolor{lightgreen}{RGB}{190, 230, 190}

\title{Probing Language Models on Their Knowledge Source}

\author{
    \textbf{Zineddine Tighidet\textsuperscript{1, 2}},
    \textbf{Andrea Mogini\textsuperscript{1}},
    \textbf{Jiali Mei\textsuperscript{1}},
    \textbf{Benjamin Piwowarski\textsuperscript{2}},
    \\
    \textbf{Patrick Gallinari\textsuperscript{2, 3}}
    \\
    \\
    \textsuperscript{1}BNP Paribas, Paris, France
    \\
    \textsuperscript{2}Sorbonne Université, CNRS, ISIR, F-75005 Paris, France
    \\
    \textsuperscript{3}Criteo AI Lab, Paris, France
    \\
    \\
    \normalsize{
        \textit{firstname.lastname@}\{\textit{isir.upmc.fr, bnpparibas.com}\}
    }
}

\definecolor{beigelight}{rgb}{1.0, 0.9, 0.90}

\begin{document}
\maketitle
\begin{abstract}

Large Language Models (LLMs) often encounter conflicts between their learned, internal (parametric knowledge, PK) and external knowledge provided during inference (contextual knowledge, CK). Understanding how LLMs models prioritize one knowledge source over the other remains a challenge. In this paper, we propose a novel probing framework to explore the mechanisms governing the selection between PK and CK in LLMs. Using controlled prompts designed to contradict the model's PK, we demonstrate that specific model activations are indicative of the knowledge source employed. We evaluate this framework on various LLMs of different sizes and demonstrate that mid-layer activations, particularly those related to relations in the input, are crucial in predicting knowledge source selection, paving the way for more reliable models capable of handling knowledge conflicts effectively.
\end{abstract}

\begin{figure}[t]
    \centering
    \includegraphics[width=1\columnwidth]{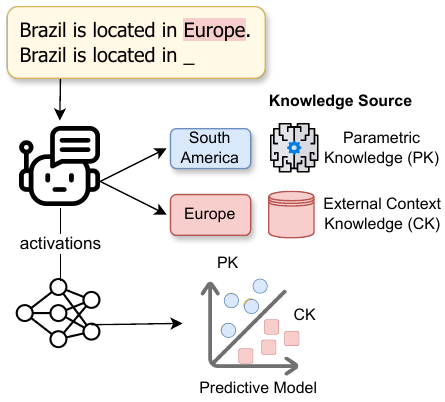}
    \caption{Illustration of our method for probing knowledge sources in LLMs. We present the model with a prompt containing contradictory information to its learned knowledge to test whether it uses parametric knowledge (PK) or contextual knowledge (CK). The resulting activations are used to train a classifier to distinguish between PK and CK.}
    \label{fig:schema_methodo}
\end{figure}

\section{Introduction}
Large Language Models (LLMs) have demonstrated remarkable proficiency in memorizing and retrieving massive amounts of information. Despite these strengths, LLMs often struggle when exposed to novel information not seen during training \cite{ovadia2019trust} or when there is a conflict between their \textbf{parametric knowledge (PK)} and the \textbf{context knowledge (CK)} provided at inference \cite{xie2024adaptive}. Such discrepancies can lead to erroneous outputs, a phenomenon that remains a significant challenge in LLMs applications \cite{Ji_2023}. While several approaches, such as reinforcement learning and retrieval-augmented generation, have been proposed to mitigate these issues \cite{ziegler2020finetuning, lewis2021retrievalaugmented}, the mechanisms by which LLMs select and prioritize knowledge sources are not well understood, suggesting a gap in current research methodologies.

This paper explores the internal dynamics of LLMs, and more precisely decoder-only layers, focusing on their decision-making processes regarding the use of CK versus PK. By prompting the LLM in a way that contradicts its PK, we probe the model's knowledge-sourcing behaviors. By training a linear classifier on model activations, our experiments reveal that certain activations correlate with determining whether context or parametric knowledge predominates in the generated outputs.

In this paper, we make the following key findings and contributions:

\begin{itemize}
    \item We define a framework that characterizes the source of knowledge used by the model to generate its outputs -- Sections \ref{sec:methodo} and \ref{sec:probing}. To facilitate further research and validation of our findings, we make our framework publicly available on GitHub\footnote{Link to the code and dataset: \url{https://github.com/Zineddine-Tighidet/knowledge-probing-framework}}.
    \item Specific activations are indicative of the knowledge source: by applying our framework to LLMs of different sizes, we establish that specific activations correlate with the model's use of contextual or parametric knowledge. 
\end{itemize}

\section{Related Work}

The understanding of the mechanisms and knowledge localization within transformers has progressed through various studies. On the one hand, some work investigated the PK-based outputs (factual setting) \cite{meng2023locating, geva2021transformer, geva2023dissecting, alkhamissi2022review, heinzerling-inui-2021-language}. These works hypothesized that LLMs store parametric information in the Multi-Layer Perceptron (MLP), which acts as a key-value memory, subsequently accessed by the Multi-Head Self-Attention (MHSA) mechanisms. On the other hand, other studies focused on the CK-based outputs and concluded that processing CK, as opposed to PK, is not specifically localized in the LLM's parameters \cite{monea2024glitch}. 

\citet{yu-etal-2023-characterizing} employed an attribution method \cite{wang2022interpretability, belrose2023eliciting} to identify the most influential attention heads responsible for generating PK and CK outputs, and subsequently adjusted the weights of these heads to modify the source of knowledge. 
Their work however focuses exclusively on knowledge specific to capital cities and relies on causal tracing, which is costly to compute. 

In contrast, our work utilizes a probing approach that uses a classifier on the LLM activations to identify the source of knowledge, leveraging the insights from previous research on the respective roles of MLPs and MHSAs in the inference process. We extend the scope of \citet{yu-etal-2023-characterizing} by incorporating a dataset with a broader range of knowledge categories~(\texttt{ParaRel} \cite{elazar2021measuring}), moving beyond just capital cities. 

\section{Methodology}
\label{sec:methodo}

We aim to show that specific activations correlate with the used knowledge source, parametric or context knowledge.
%
In order to probe LLMs, we construct prompts that are composed of inputs representing information about a subject $s$ that contradicts what the model learned during training, followed by a query about the same subject (see Figure \ref{fig:schema_methodo}). If the model answers according to the prompt, then it uses context knowledge. On the other hand, if the model answers according to what it learned, then it is using its parametric knowledge. In the following two sections, we define more formally PK and CK.

\subsection{Parametric Knowledge (PK)}
\label{subsec:pk}
We consider the parametric knowledge (PK) to be the information that the model learned during training. More specifically, we restrict this PK by using a knowledge base $KB=\{(s, r, o)\}$, i.e. a set of (subject, relation, object) triplets from the ParaRel dataset \cite{elazar2021measuring}. We then define $PK$ to be the set of objects that are generated by a LLM:
\begin{flalign}
    \label{formula:c}
    PK = \{ (s, r, o^{'}) \mid \exists o\ \mathrm{s.t.}\ (s, r, o) \in KB \nonumber \\
    \wedge \text{ } o^{'} = G_{\theta}(q(s, r)) \}
\end{flalign}
where $G_{\theta}$ is an LLM; $q(s, r)$ is a prompt in natural language corresponding to a subject-relation pair $(s, r)$; $o^{'}$ is the output of $G_{\theta}$ given the query prompt (e.g. \textit{"Brazil is located in the continent of \_"}).

Note that we use this method to define PK because we do not have access to the training data of LLMs in general, and, more importantly, we are interested in what the LLM infers by itself.
If $o = G_{\theta}(q(s, r))$, that is, the object $o$ was generated by the model after providing an input query $q(s, r)$, we can conclude that the model learned to associate the object $o$ with the subject $s$ with the relation $r$ during training. Note also that, unlike previous work \cite{meng2023locating, yu-etal-2023-characterizing}, even when $o$ is factually incorrect (e.g. \textit{"Paris is the capital of Italy"}), we still consider it in our study as our only interest is the parametric knowledge and not the external world factual truth\footnote{This behavior happens when the subjects are unpopular and the LLM was not trained on enough examples. We discuss this further in Section \ref{subsec:freq_source}.}.

\begin{figure*}[t]
    \centering
    \includegraphics[width=6.3in]{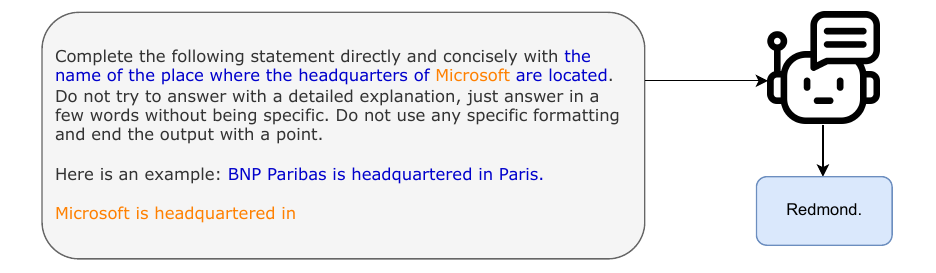}
    \caption{Example of the template used to generate the parametric knowledge dataset. The blue text is proper to the relation and the orange is specific to a subject-relation example in the ParaRel dataset \cite{elazar2021measuring}.}
    \label{fig:parametric_prompt}
\end{figure*}

\subsubsection{Knowledge Base (ParaRel)}
We extend the ParaRel dataset \cite{elazar2021measuring} for constructing a parametric knowledge base. ParaRel dataset consists of triplets, each composed of a subject, a relation, and an object. Table \ref{table:raw_pararel} illustrates a sample of the raw ParaRel dataset.

While the majority of the triplets adhere to the subject-relation-object structure, some deviate from this format. To ensure consistency, a pre-processing step was applied on the raw ParaRel dataset using Mistral-Large\footnote{\url{https://mistral.ai/news/mistral-large/}}. Specifically, the goal was to transform triplets where the subject precedes the relation (e.g., \textit{"The official language of France is French."}) into triplets where the subject is placed directly before the relation (e.g., \textit{"France's official language is French."}). We selected Mistral-Large because it is open-weight, enabling reproducibility, and its capabilities are very close to those of GPT-4.

\begin{table*}[htbp]
\small
\centering
\begin{tabular}{lllll}
    \toprule
    \textbf{subject} & \textbf{rel-lemma} & \textbf{object} & \textbf{query} \\ \midrule
    Newport County A.F.C. & is-headquarter & Newport & Newport County A.F.C. is headquartered in \\ 
    Norway & capital-city-of & Oslo & Norway's capital city, \\ 
    WWE & is-headquarter & Stamford & WWE is headquartered in \\ 
    Princeton University Press & is-headquarter & Princeton & Princeton University Press is headquartered in \\ 
    Internet censorship & is-subclass & censorship & Internet censorship is a subclass of \\ 
    McMurdo Station & part-of-continent & Antarctica & McMurdo Station is a part of the continent of \\ 
    Windows Update & product-manufacture-by & Microsoft & Windows Update, a product manufactured by \\ 
    Nintendo & located-in & Kyoto & The headquarter of Nintendo is located in \\ 
    Microsoft Windows SDK & product-manufacture-by & Microsoft & Microsoft Windows SDK, a product manufactured by \\ 
    Harare & capital-of & Zimbabwe & Harare, the capital of \\ 
    \bottomrule
\end{tabular}
\caption{A sample of the raw ParaRel dataset \cite{elazar2021measuring}}
\label{table:raw_pararel}
\end{table*}

\subsubsection{Parametric Knowledge Query Format}
To guide the studied LLMs towards generating parametric knowledge objects that are coherent with the relation and to help specifying the type of object that is expected when there are multiple possible answers (for example in \textit{"Napoleon passed away in"} the LLM can generate the place of death \textit{"Longwood"} or the year of death \textit{"1821"}) we propose to use a template prompt that is illustrated in Figure~\ref{fig:parametric_prompt}. The prompt specifies the requested type of object with a brief description as well as an example (one-shot learning) to help the LLM understand the kind of object that is intended (illustrated in blue in Figure~\ref{fig:parametric_prompt}). The description and example were manually created for each relation. The prompt also tries to guide the LLM towards generating a concise output as these models tend to give a long explanation that is irrelevant in our study (e.g. \textit{Amazon is headquartered in the city of Seattle where Starbucks is also headquartered...}).

\subsubsection{Subject/Object Bias}
\label{subsec:sub_obj_bias}
The subject can sometimes provide relevant information about the object which can bias our definition of parametric knowledge (e.g. \textit{Princeton University Press is located in Princeton.} or \textit{Niger shares the border with Nigeria}). To avoid this, we removed examples where the subject is similar to the object, utilizing the Jaro-Winkler string distance \cite{wiki:jw} with a threshold empirically fixed at 0.8. This method is advantageous for our dataset, as it assigns closer distances to subjects with the same prefix as the objects, which is common in cases like \textit{"Croatia's official language is Croatian"} where \textit{"Croatia"} and \textit{"Croatian"} have the same prefix.

\subsection{Context Knowledge (CK)}
\label{sec:ck}

In our framework, we perturb the LLM by providing a CK that contradicts the PK, which we name \emph{counter-PK} and denote $\overline{PK}$.
It is challenging to test what the model does not know \cite{yin-etal-2023-large}. One way to build these inputs is to contradict what the model learned during training by taking $(s, r, o) \in PK$ and replacing $o$ with another object $\bar{o} \in O_{r}$ that shares the same relation $r$ to keep semantic consistency~(e.g. \textit{"Elvis Presley is a citizen of Japan"}, here we replaced \textit{"the USA"} with a country name: \textit{"Japan"}).
More specifically, the set of tuples $\overline{PK}$ that represents the counter-PK is defined as follows:

\begin{align}
    \label{formula:c_bar}
    \overline{PK} &= \bigcup_{(s, r, o) \in PK} \texttt{Counter-PK}_{k}(s, r, o)
\end{align}

where:

\begin{multline}
     \texttt{Counter-PK}_{k}(s, r, o) = \quad \{ (s, r, o, \bar{o}) \mid \bar{o} \in O_{r} \wedge \\
       \bar{o} \neq o \wedge \textit{rank}_{\theta}(\bar{o}\mid s, r) \leq k \} 
\end{multline}

\noindent where $k$ is the number of counter-knowledge triplets per triplet $(s, r, o)$ in PK; $\textit{rank}_{\theta}(o\mid s, r)$ is the rank of $\bar{o}$ among the $O_r$ ordered by the increasing probability $p(\hat{o} \mid q(s, r))$ of the LLM to generate an object $\hat{o} \in O_{r}$ given the prompt $q(s, r)$. We also make sure that the model has not learned the $(s, r, \bar{o})$ association by considering the objects $\hat{o}$ with the $k$ lowest ranks ($\textit{rank}_{\theta} \leq k$) -- indicating that the LLM is very unlikely to use its parametric knowledge to generate $\bar{o}$. 

\begin{figure}[t]
    \centering
    \includegraphics[width=1.1\columnwidth]{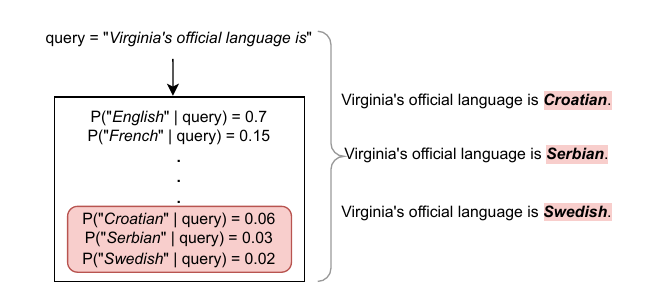} 
    \caption{Example of 3 counter-knowledge objects that were associated to a parametric knowledge element. The probability distribution is ranked in an descendant order and we selected the objects with the lowerst probabilities.}
    \label{fig:counter_knowledge_examples}
\end{figure}

Figure~\ref{fig:counter_knowledge_examples} illustrates the counter-knowledge objects that were generated by Phi-1.5 for a parametric knowledge example.

\subsection{Models}
We consider decoder-only Transformer models. Between layer $l$ and $l-1$, the hidden state $X^{(l-1)}$ is updated by:
\begin{align}
    \label{formula:resid}
    X^{(l)} = \gamma\text{(}X^{(l-1)} + A^{(l)})\text{)} + M^{(l)}
\end{align}

\noindent where $A^{(l)}$ and $M^{(l)}$ are the outputs of the MHSA and MLP modules respectively, and $\gamma$ is a non-linearity.

The MLP module is a two-layer neural network parameterized by matrices $W_{\textit{mlp}}^{(l)} \in \mathbb{R}^{d \times d_{\textit{mlp}}}$ and $W_{\textit{proj}}^{(l)} \in \mathbb{R}^{d_{\textit{mlp}} \times d}$: 
\begin{align}
    \label{formula:mlp}
    M^{(l)} = \sigma (X^{(l)}_{mlp} W_{\textit{mlp}}^{(l)}) W_{\textit{proj}}^{(l)} \in \mathbb{R}^{n \times d}
\end{align}

\noindent where $\sigma$ is a non-linearity function (e.g. GeLU) and $X^{(l)}_{mlp}$ is the input of the MLP. We refer the reader to \citet{NIPS2017_3f5ee243} for more details on the architecture.

In our probing set-up (Section \ref{sec:probing}), we use the following activations: $\sigma (X^{(l)}_{mlp} W_{\textit{mlp}}^{(l)})$ the first layer of the MLP (referred as MLP-L1 in this paper), $\sigma (X^{(l)}_{mlp} W_{\textit{mlp}}^{(l)}) W_{\textit{proj}}^{(l)}$ the output of the MLP (i.e. second layer, referred as MLP-L2 in this paper), and $A^{(l)}$ the output of the MHSA. We consider the first and second MLP layers activations, based on \citet{geva2021transformer} work, and also 
 the MHSA activations as the attentions play a crucial role in information selection from the MLP memory \cite{geva2023dissecting}.

We evaluate our method on several LLMs with different sizes: \texttt{Phi-1.5} with 1.3B parameters \cite{li2023textbooks}, \texttt{Pythia-1.4B} with 1.4B parameters \cite{biderman2023pythia}, \texttt{Mistral-7B} with 7B parameters \cite{jiang2023mistral}, and \texttt{Llama3-8B} with 8B parameters \cite{llama3modelcard}. Table \ref{tab:models_characteristics} gives characteristics about the LLMs' modules dimensions.

\begin{table}[h]
\centering
\begin{tabular}{|c|c|c|c|}
\hline
\textbf{Model} & \textbf{MLP-L2} & \textbf{MLP-L1} & \textbf{MHSA} \\
\hline
\texttt{Phi-1.5}   & 2048 & 8192 & 2048 \\
\texttt{Pythia-1.4B} & 2048 & 8192  & 2048  \\
\texttt{Llama3-8B}  & 4096 & 14336 & 4096 \\
\texttt{Mistral-7B} & 4096 & 14336 & 4096 \\
\hline
\end{tabular}
\caption{Activation dimensions for Phi-1.5, Pythia-1.4B, Llama3-8B and Mistral-7B for the different considered modules (MLP-L2, MLP-L1 and MHSA)}
\label{tab:models_characteristics}
\end{table}

\textbf{Decoding strategy} As the generated sequences are short, we use a greedy decoding strategy and limit the number of generated tokens to 10.

\begin{table*}[htbp]
\small
\centering
\begin{tabular}{lp{9cm}c p{8cm}} 
    \toprule
    \textbf{Relation Group ID} & \textbf{Relations} & \textbf{\#Examples} \\ \midrule
    \rowcolor{lightgray} geographic-geopolitic-language & \textit{is-headquarter, located-in, headquarters-in, locate, share-border, is-twin-city-of, located, border-with, is-located, work-in-area, which-is-located, capital-city-of, part-of-continent, capital-of, headquarter, belong-to-continent, based-in, is-citizen-of, that-originate-in, originate-in, is-in, found-in, share-common-border, is-native-to, is-originally-from, pass-away-in, born-in, hold-citizenship-of, have-citizenship-of, citizen-of, start-in, formulate-in, legal-term, tie-diplomatic-relations, maintains-diplomatic-relations, have-diplomatic-relations, native, mother-tongue, original-language-is, the-official-language, communicate} & 2815 \\ 
    \midrule
    corporate-products-employment & \textit{product-manufacture-by, develop-by, owned-by, product-develope-by, product-release-by, create-by, product-of, produce-by, owner, is-product-of, is-part-of, who-works-for, employed-by, who-employed-by, works-for, work-in-field, profession-is, found-employment} & 1217 \\ 
    \midrule
    \rowcolor{lightgray} media & \textit{premiere-on, to-debut-on, air-on-originally, debut-on} & 128 \\ 
    \midrule
    religion & \textit{official-religion} & 249 \\ 
    \midrule
    \rowcolor{lightgray} hierarchy & \textit{is-subclass} & 183 \\ 
    \midrule
    naming-reference & \textit{is-call-after, is-name-after, is-name-for} & 6 \\ 
    \midrule
    \rowcolor{lightgray} occupy-position & \textit{play-in-position, who-holds} & 77 \\ 
    \midrule
    play-instrument & \textit{play-the} & 13 \\ 
    \bottomrule
\end{tabular}
\caption{All the relation groups with their corresponding relations and number of examples.}
\label{table:relation_groups}
\end{table*}

\section{Probing Set-up}
\label{sec:probing}
To build our probing dataset, we associate each tuple
$(s, r, o, \bar{o})\in\overline{PK}$ with a prompt $prompt(s, r, \bar{o})$
that corresponds to a natural language statement of $(s,r,\bar o)$ 
followed by $q(s,r)$ (see Figure \ref{fig:schema_methodo}).
Each prompt is associated with a label among CK, PK, and ND, where
\textbf{CK} if $G_{\theta}(prompt(s, r, \bar{o})) = \bar{o}$, \textbf{PK} if $G_{\theta}(prompt(s, r, \bar{o})) = o$, and with ND (Not Defined) otherwise. In this work, we discard tuples associated with ND.

We specifically probe the activations $\bar{o}$ of the object, $s_{q}$ of the subject in the query, and $r_{q}$ the relation in the query. As each of these elements may have multiple tokens, we use their last tokens as their representative (e.g. for \textit{"Washington"} $\rightarrow$ \textit{["Wash", "inghton"]}, we consider the activations of the token \textit{"inghton"}). The fact that this token representation summarizes the entity is intuitively true for decoder-only models and has been experimentally validated in~\cite{meng2023locating, geva2023dissecting}.

Note that our first probe targets $\bar{o}$ as this is where the knowledge conflict starts (e.g. \textit{Bill Gates is the founder of \textbf{\underline{Apple}($\bar{o}$)\textbf{. \underline{Bill Gates}}($s_q$)\textbf{ \underline{is the founder of}}($r_q$)} \_}).

\subsection{Control experiments}
\label{subsec:control}
We also probe the activations of the first token to measure how much of the prediction can be attributed to the subject representation itself. Since the knowledge perturbation starts with the first object token, the first token activations should not indicate the knowledge source. For instance, in \textit{Paris is located in Italy} the representation of the first token (\textit{Paris}) should not contain information about the knowledge source as the perturbation starts at \textit{Italy}.

\subsection{Relation Groups}
To avoid syntactic and semantic biases related to the type of relation when training a classifier, we grouped the relations that are similar into relation groups. The relation groups are illustrated in Table~\ref{table:relation_groups}.


\subsection{Evaluation}
\label{sec:eval}
We use each relation group as a test set and train on the rest of the relation groups. We make sure that the train and test sets do not share similar subjects and objects to avoid biases related to the syntax or the nature of the relation and subject. We ensure the train set is balanced (equal number of CK and PK), as current LLMs are more likely to use context information (CK) than their parametric knowledge~\citet{xie2024adaptive}. This is illustrated by Figure \ref{fig:output_labels_per_model} (and Figure \ref{fig:rel_proportions} in appendix for a breakdown by relation), where we can see that the considered LLMs mostly generate CK-based outputs.

We also ensure that the test set is balanced so we can use the success rate (accuracy) as the main metric –– with 50\% being the performance of a random classifier. We compute the success rate $p_i$ for each group of relations. As $p_i$ follows a binomial distribution, we used a binomial proportion confidence interval to compute the weighted standard error (WSE -- see formula \ref{formula:error_propagation}) around the average success rate (see formula \ref{formula:averaged_P}) with a 95\% confidence interval to assess the significance of the resulting classification scores for each layer and token. We used the following formula in order to propagate the errors across the relation groups:
\begin{align}
    \label{formula:error_propagation}
    \text{WSE} = \sqrt{\sum_{i=1}^{G} \left( \frac{n_i}{N} \times \text{SE}_{i} \right)^2}
\end{align}

Where $\text{SE}_i$ is the standard error for the $i^{th}$ relation group, defined as:
\begin{align}
    \label{formula:se_i}
    \text{SE}_i = \sqrt{\frac{p_i \times (1-p_i)}{n_i}}
\end{align}

\noindent $G=8$ is the number of relation groups; $n_i$ is the number of test examples for the $i^{th}$ relation group; $N$ is the total number of test examples across all the relation groups.

The error bars are finally computed using a $z$-score of 1.96 for a confidence interval of 95\%:
\begin{align}
    \text{Error Bars} = [P - 1.96 \times \text{WSE}, P + 1.96 \times \text{WSE}]
\end{align}

Where $P$ is the average success rate across all the relation groups:
\begin{align}
    \label{formula:averaged_P}
    P = \frac{\sum_{i=1}^{G}n_i \times p_i}{N}
\end{align}

Figure \ref{fig:success_rate} presents the success rates for classifiers trained on activations from object, subject, and relation tokens, with the first token used as a control (see Section \ref{subsec:control} for more details on the control experiment.) Results are reported for Mistral-7B, Phi-1.5, Llama3-8B, and Pythia-1.4B. Solid lines represent the average success rates across relation groups, while shaded areas denote the weighted standard error with a 95\% confidence interval.

\begin{figure}[t]
    \centering
    \includegraphics[width=1\columnwidth]{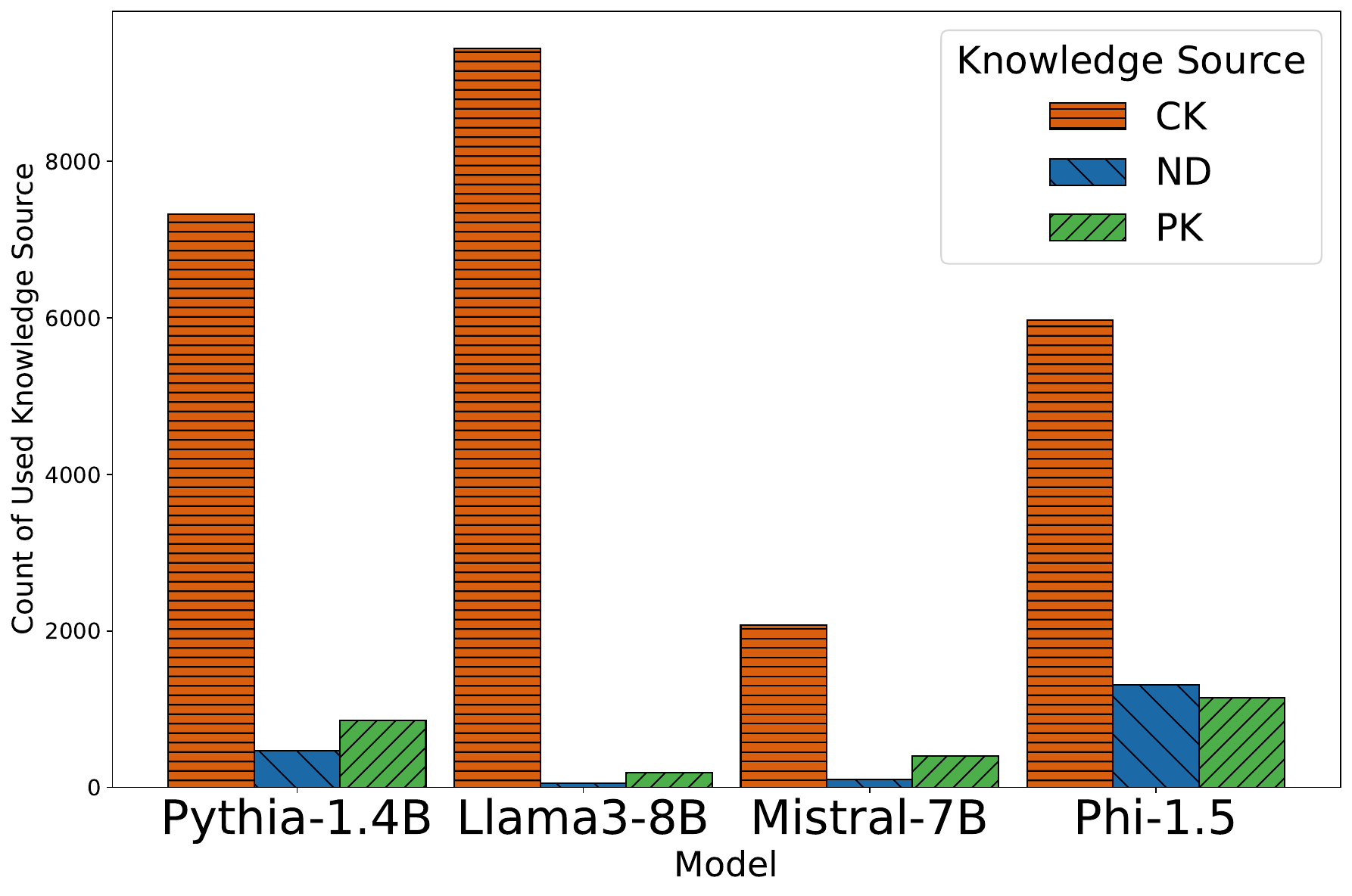} 
    \caption{Count of used knowledge sources by each model (CK, PK, and ND). ND refers to outputs where the knowledge source is not defined.}
    \label{fig:output_labels_per_model}
\end{figure}

\section{Results and Discussion}
\label{sec:results}
\begin{figure*}[h!]
    \centering
    \includegraphics[width=6.2in]{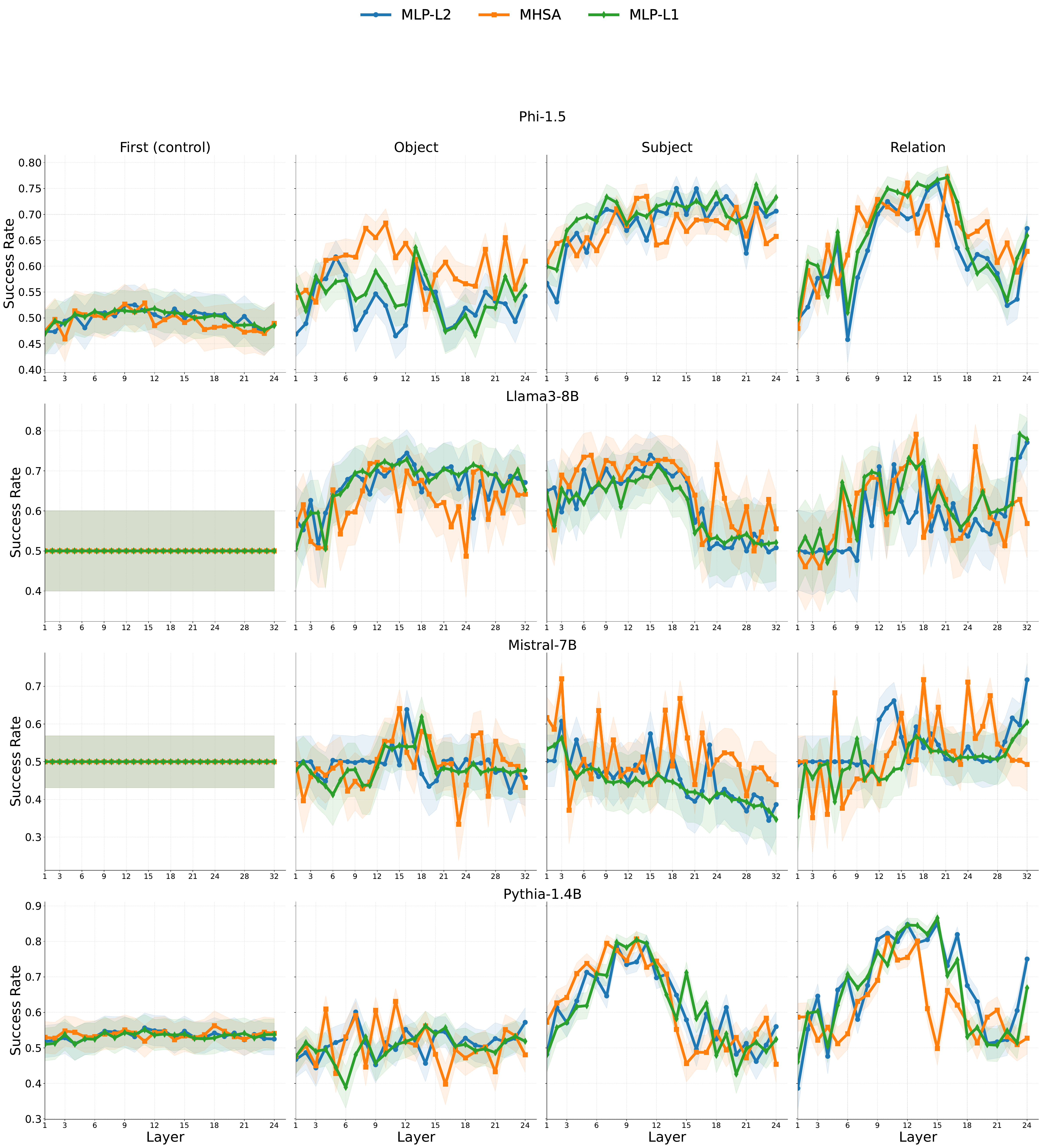}
    \caption{Performance of the linear classifier in identifying knowledge sources across different layers and modules (MLP-L2, MLP-L1, MHSA). The plots show success rates for classifiers trained on activations from object, subject, and relation tokens, with the first token used as a control (see Section \ref{subsec:control} for more details on the control experiment.) Results are reported for the Mistral-7B, Phi-1.5, Llama3-8B, and Pythia-1.4B models. Solid lines represent the average success rates across relation groups, while shaded areas denote the weighted standard error with a 95\% confidence interval. See Section \ref{sec:eval} for further details on the evaluation methodology.}
    \label{fig:success_rate}
\end{figure*}

In Figure \ref{fig:success_rate}, we can first observe that the features linked to $\bar o$, the subject $s_q$ and the relation $r_q$ exhibit a correlation with the used knowledge source for MLP-L2, MLP-L1, and MHSA activations.

Our framework successfully transfer the learned knowledge source patterns from one relation group to another, generalizing well from one group to another. The most predictive features are those of $r_q$, i.e. the relation token in the query. On certain layers, the success rate increases significantly, reaching 87\% for Pythia-1.4b on the 15th layer at the relation token position.

This finding is consistent with prior research, which indicates that LLMs primarily store knowledge in the MLPs \cite{meng2023locating, geva2021transformer}. Moreover, it supports \citet{geva2023dissecting}'s insights on the information extraction process, where the relation token retrieves attributes from $s_q$'s MLPs using the MHSA (a process referred to as \textit{Attribute Extraction}).

Additionally, it is noteworthy that the knowledge source can be detected directly starting from the perturbing object $\bar{o}$. This shows that detecting a potentially harmful conflict knowledge statement is possible early in the LLM inference process. MHSA activations are less connected to the used knowledge source than MLP-L2 and MLP-L1 activations.

The results of the control experiments conducted on the first token of the input indicate that the learned patterns in the object, subject, and relation are not arbitrary. The success rates of all LLMs for the first token appear to be random (about 0.5).

Finally, compared to \cite{yu-etal-2023-characterizing}, we show in this work that it is possible to predict the knowledge source based on the sole activations of an LLM, and, even more importantly, that we predict this for multiple relations rather than being limited to a single relation.

\begin{table*}[htbp]
\small
\centering
\begin{tabular}{>{\raggedright\arraybackslash}p{9cm}ccc} 
    \toprule
    \textbf{Input Prompt} & \textbf{Knowledge Source} & \textbf{PK Object} & \textbf{Model} \\
    \midrule
    \rowcolor{lightblue} \textit{Harney County has its capital city in \underline{Taiwan}. Harney County has its capital city in \textbf{Burns.}} & ND & Oregon & Llama3-8B \\ 
    \midrule
    \rowcolor{lightblue} \textit{Lisa Appignanesi has citizenship of \underline{Finland}. Lisa Appignanesi has citizenship of \textbf{France.}} & ND & the UK & Llama3-8B \\ 
    \midrule
    \rowcolor{lightblue} \textit{Craiova is located in the continent of \underline{India}. Craiova is located in the continent of \textbf{Romania.}} & ND & \textit{Europe} & Pythia-1.4B \\ 
    \midrule
    \rowcolor{lightorange} \textit{The Kingdom of Hungary had its capital as \underline{Connecticut}. The Kingdom of Hungary had its capital as \textbf{Connecticut.}} & CK & \textit{Budapest} & Mistral-7B \\ 
    \midrule
    \rowcolor{lightorange} \textit{The Wii U system software is a product that was manufactured by \underline{Square}. The Wii U system software is a product that was manufactured by \textbf{Square.}} & CK & \textit{Nintendo} & Llama3-8B \\ 
    \midrule
    \rowcolor{lightorange} \textit{The Centers for Disease Control and Prevention is headquartered in \underline{Lyon}. The Centers for Disease Control and Prevention is headquartered in \textbf{Lyon.}} & CK & \textit{Atlanta} & Llama3-8B \\ 
    \midrule
    \rowcolor{lightgreen} \textit{Harare is the capital city of \underline{Florida}. Harare is the capital city of \textbf{Zimbabwe.}} & PK & \textit{Zimbabwe} & Pythia-1.4B \\ 
    \midrule
    \rowcolor{lightgreen} \textit{Goodreads is owned by \underline{Microsoft}. Goodreads is owned by \textbf{Amazon.}} & PK & \textit{Amazon} & Phi-1.5 \\ 
    \midrule
    \rowcolor{lightgreen} \textit{OneDrive is owned by \underline{Toyota}. OneDrive is owned by \textbf{Microsoft.}} & PK & \textit{Microsoft} & Mistral-7B \\ 
    \bottomrule
\end{tabular}
\caption{Examples of final probing prompts, including their knowledge source, the LLM, and the corresponding parametric knowledge (PK) object. Bold text indicates the generated object, while underlined text represents the counter-knowledge object.}
\label{table:prompting_sample}
\end{table*}

\begin{figure}[t]
    \centering
    \includegraphics[width=1\columnwidth]{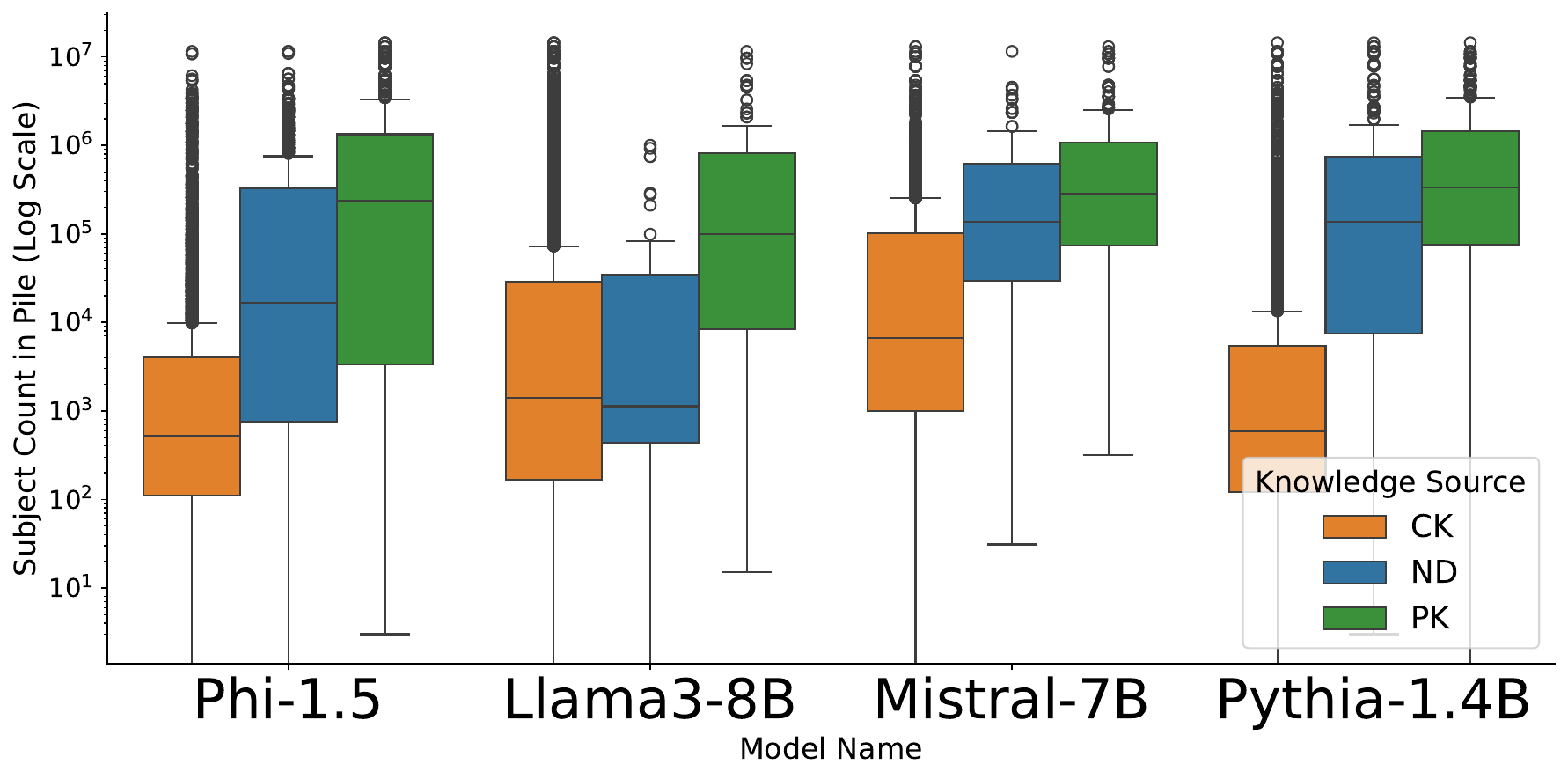}
    \caption{Subject frequency in the training dataset (The Pile) for CK, PK, and ND outputs. We use The Pile as an approximation of what the LLMs might have learned except for Pythia-1.4B for which it is the actual training data.}
    \label{fig:subj_count}
\end{figure}

\section{Subject frequency Vs. Knowledge Source}
\label{subsec:freq_source}
To understand what makes an LLM select the CK object over the PK object, we observed the subject frequency in The Pile corpus \cite{gao2020pile800gbdatasetdiverse} for CK, PK, and ND outputs as illustrated in Figure \ref{fig:subj_count} -- We use The Pile as an approximation of what the LLMs might have learned except for Pythia-1.4B for which it is the training data. We used the infini-gram API made available by \citet{Liu2024InfiniGram} in order to get the frequencies. A Mann-Whitney U test reveals that the subject frequency distribution for PK outputs is significantly higher than for CK and ND outputs, except in the case of Pythia-1.4B, where PK is only higher than CK but not ND. This suggests that as a model gains more knowledge about a subject, it becomes more likely to select PK over CK objects.

\section{Probing Dataset Examples}

Table \ref{table:prompting_sample} illustrates some examples of the final probing prompts with their knowledge source, the LLM, and the corresponding PK object.

\section{Conclusion}

In this study, we introduced a novel probing framework to investigate if we can detect when LLMs switch from PK to CK. Our findings reveal that specific model activations are significantly correlated with the used knowledge source with a success rate reaching 87\% for Pythia-1.4b. Additionally, our framework is able to transfer the learned knowledge source patterns from one relation group to another. This opens the door for future work investigating the mechanism at play when using CK or PK, and finally to building models that can better control this behavior.

\section{Limitations}

Our current framework is designed to probe LLMs by introducing contradictions to their learned knowledge, effectively identifying the source of knowledge. However, this controlled experimental setting does not account for many other situations, e.g. where the knowledge remains unperturbed. Future work should extend the framework to handle cases where both the parametric knowledge (PK) and the contextual knowledge (CK) are consistent or not related, providing a more comprehensive understanding of LLM behavior. Additionally, our study primarily measures the correlation between specific activations and the use of PK or CK, which, while providing valuable insights, does not establish an explanation of the underlying process. Further research is needed to uncover the underlying mechanisms that govern knowledge source selection in LLMs, possibly through experimental designs that manipulate specific model parameters or activations to observe resulting behavioral changes.

It might also be interesting to employ a variety of prompt structures to mitigate biases associated with the conventional subject-relation-object format. Exploring alternative combinations, such as relation-subject-object (e.g., \textit{The official language of Italy is Italian}), could yield valuable insights.

\section{Ethical Considerations}

Our probing framework of LLMs for their knowledge-sourcing behaviors only uses publicly available, non-personal datasets to ensure privacy and security. We recognize the potential for misuse of our findings. The insights derived from our research could be exploited to generate misleading information or make the models more susceptible to adversarial attacks. Therefore, we emphasize the importance of the ethical application of our work. Researchers and practitioners must implement robust safeguards to prevent the misuse of these technologies and ensure they are used to benefit society. Developing and deploying robust security measures is essential to protect against these vulnerabilities and maintain the integrity of information generated by LLMs. While we recognize inherent biases in LLMs, our commitment to transparency is demonstrated through the public release of our framework, facilitating reproducibility and further research.

\section{Acknowledgements}
We would like to thank BNP Paribas and the French National Association for Research and Technology (ANRT) who founded this project under the CIFRE program (2023/1673). We also thank Etienne Boisseau for his help on this paper.

\bibliography{custom}

\appendix
\label{sec:appendix}

\section{Data Characteristics}
The \texttt{ParaRel} \cite{elazar2021measuring} dataset includes 5313 unique subject-relation pairs, leading to the formation of the same number of PK triplets. After removing the examples where the subject is similar to the parametric object (see Section \ref{subsec:sub_obj_bias}) we are left with approximately 3600 examples depending on the LLMs' parametric knowledge. We take $k=3$ for $\texttt{Counter-PK}_{k}$ which gives approximately counter-PK 10k triplets. After under-sampling, we are left with approximately 2000 balanced prompts depending on the LLM.

\section{Data Undersampling Seed Impact}
To examine the impact of random seed selection in undersampling for balanced CK and PK classes, we conducted an experiment with various seeds to determine if seed choice influenced model performance. Changing the seed for the undersampling of the majority class introduces significant variations in the success rate of our classifiers. This effect can be explained by the fact that the minority class (PK) has fewer samples than the majority class (CK), meaning there are very few CK examples in common between the datasets generated by two different seeds. In some cases, we observed a standard deviation of up to 11 points of accuracy in the success rate when varying the seed. However, the results stayed consistent with our conclusions across all choices of seeds. 

\section{Hardware and Software}

Text generation tasks were performed using A100 GPUs, each equipped with 80 GB of memory. The process of generating the outputs spanned around 100 GPU hours. Our framework was constructed utilizing PyTorch \cite{paszke2019pytorch} and the HuggingFace Transformers library \cite{wolf2020huggingfaces}.

\section{License}

\textbf{Model weights.} Llama3-8B weights are released under the license available at \url{https://llama.meta.com/llama3/license/}. Mistral-7B and Pythia-1.4B weights are released under an Apache 2.0 license. Mistral-Large weights are released under the licence available at \url{https://mistral.ai/licenses/MRL-0.1.md}. Phi-1.5 weights are released under a MIT license.

\textbf{Data.} The ParaRel dataset we used is released under a MIT License.

\begin{figure*}[t]
    \centering
    \includegraphics[width=6.7in]{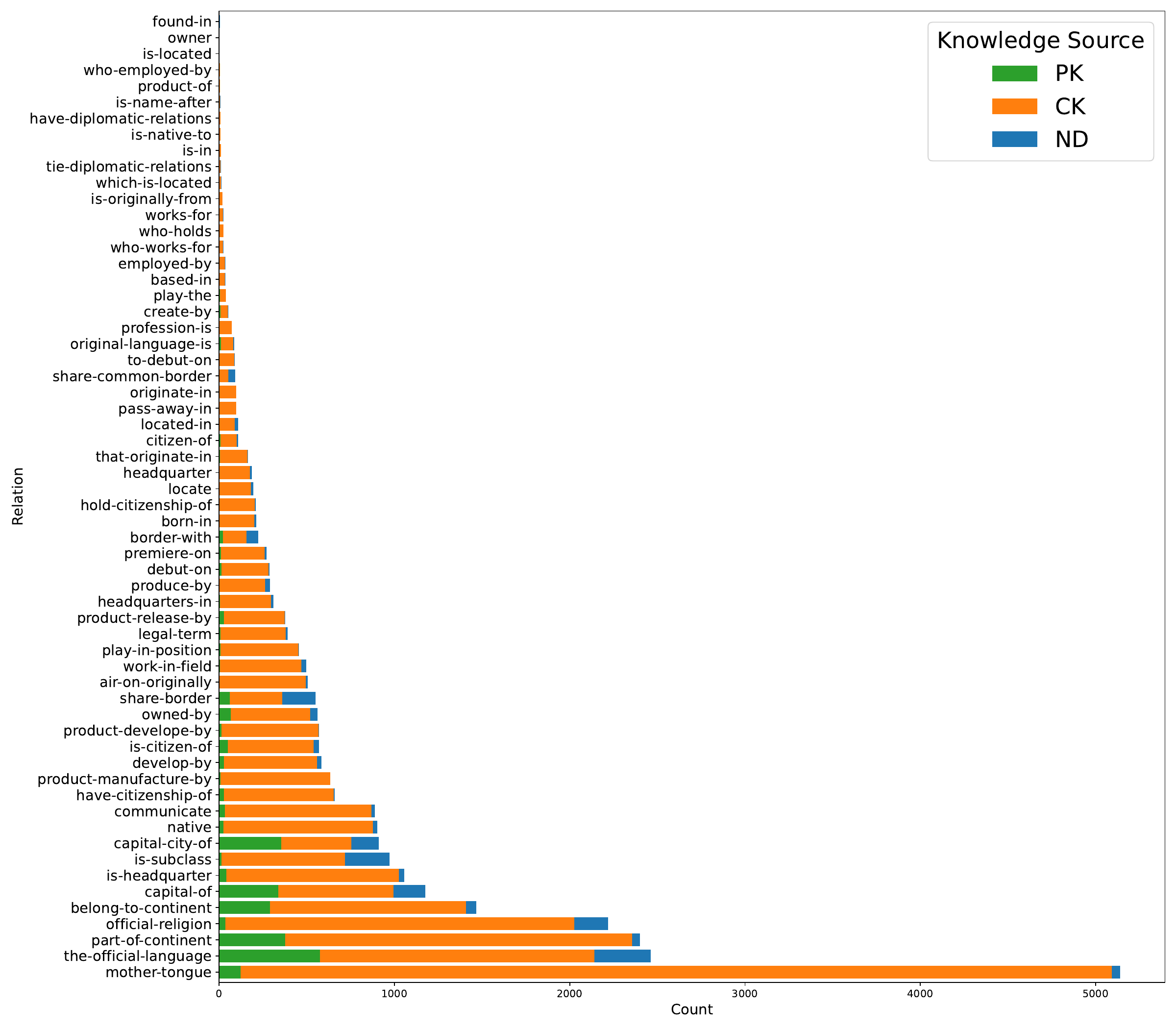}
    \caption{All the considered relations with the number of outputs that used CK (orange), PK (green), and ND (blue) sources (the counts include all the considered LLMs).}
    \label{fig:rel_proportions}
\end{figure*}

\end{document}